\documentclass[conference]{IEEEtran}
\IEEEoverridecommandlockouts
\usepackage{cite}
\usepackage{amsmath,amssymb,amsfonts}
\usepackage{algorithm}
\usepackage{algorithmic}
\usepackage{graphicx}
\usepackage{textcomp}
\usepackage{xcolor}
\def\BibTeX{{\rm B\kern-.05em{\sc i\kern-.025em b}\kern-.08em
    T\kern-.1667em\lower.7ex\hbox{E}\kern-.125emX}}

\usepackage{pifont}
\usepackage{amsmath}
\usepackage{enumitem}
\usepackage{lineno}
\usepackage{tabularx}
\usepackage{makecell}
\usepackage{subcaption}
\usepackage{makecell}
\usepackage{fancyhdr}
\usepackage{amsfonts}
\usepackage{multirow}
\usepackage{comment}
\usepackage{hyperref}
\usepackage[noabbrev]{cleveref}

\usepackage{booktabs}
\usepackage{siunitx}

\usepackage{listings}
\usepackage{xcolor}

\definecolor{dkgreen}{rgb}{0,0.6,0}
\definecolor{gray}{rgb}{0.5,0.5,0.5}
\definecolor{mauve}{rgb}{0.58,0,0.82}
\lstdefinestyle{custompython}{
  language=Python,
  keywordstyle={},               
  commentstyle=\color{green!50!black}, 
  stringstyle=\color{orange},   
  basicstyle=\ttfamily\small,
  showstringspaces=false,
  breaklines=true,
  inputencoding=utf8
}
    
\begin{document}

\title{FICHAD: Fusion of Image Context and Human-Annotated Descriptions for Multi-Modal Knowledge Graph Completion
}

\author{
\IEEEauthorblockN{Haodi Ma, Dzmitry Kasinets, Daisy Zhe Wang}
\IEEEauthorblockA{
\textit{Department of Computer and Information Science and Engineering, University of Florida}, Gainesville, USA \\
\{ma.haodi, dkasinets, daisyw\}@ufl.edu}
}

\maketitle

\begin{abstract}
Multimodal knowledge graph completion (MMKGC) aims to predict missing links in multimodal knowledge graphs (MMKGs) by leveraging structural, textual, and visual information. This is critical in real-world applications such as identifying fine-grained visual facts in art, associating damaged goods with vendors in supply chain audits, or connecting medical scans to clinical entities where language-only models often fail due to outdated or incomplete world knowledge, and vision-language models (VLMs) struggle with reasoning over structured relations.

We propose \textbf{FICHAD}, a modular framework that uses pre-trained VLMs to generate multimodal context in textual form, which includes link-aware descriptions (\textbf{LAMM-context}), entity-centric summaries, and conceptual hints for relation understanding. This context can be directly injected into existing description-based KGC models (e.g., SimKGC, CSProm-KG) without retraining or architectural changes.

Evaluated on three public MMKG benchmarks (FB15K-237-IMG, MKG-W, MKG-Y), FICHAD consistently improves performance; our best variant outperforms prior state-of-the-art MMKGC models by up to 4.6\% in MRR and 3.5\% in Hits@1 metrics. We further provide quantitative and qualitative analyses demonstrating that VLM-generated context captures relational semantics often missed by baseline methods. Our findings show that FICHAD offers a scalable and effective framework for integrating multimodal knowledge into symbolic reasoning tasks like KGC.
\end{abstract}


\section{Introduction}
Knowledge graphs (KGs) such as YAGO~\cite{mahdisoltani2014yago3}, Freebase~\cite{bollacker2008freebase}, and Wikidata~\cite{vrandevcic2014wikidata} represent facts as structured triples $(h, r, t)$, where $h$ and $t$ are entities and $r$ is a relation. They support applications such as recommendation, question answering, and retrieval-augmented generation (RAG)~\cite{sanmartin2024kg}, but are often incomplete, limiting their utility.

To address this, knowledge graph completion (KGC) aims to predict missing links. While early methods use embeddings~\cite{bordes2013translating,sun2018rotate}, recent \textit{description-based} models~\cite{wang-etal-2022-simkgc,chen-etal-2023-dipping} incorporate textual entity descriptions to improve generalization, particularly for unseen entities. However, these models are limited to textual input and cannot leverage multimodal evidence such as images.

Multimodal knowledge graphs (MMKGs) extend KGs with entity-linked visual content, enabling reasoning tasks that require both symbolic and perceptual cues. For example, identifying which plant appears in a painting like \textit{View of Arles} requires mapping visual attributes (e.g., white flowers) to structured types. In industrial applications like supply chain monitoring~\cite{infosys2022smart}, linking photos of damaged goods to shipment records demands integration of images and structured metadata—beyond the scope of language-only or vision-only models.

These use cases motivate the need for \textit{structured visual reasoning}, where symbolic relations and visual signals are jointly modeled. While vision-language models (VLMs) like BLIP~\cite{li2022blip} and QWEN-VL~\cite{Qwen-VL} excel at generating rich image captions, existing MMKGC methods~\cite{xie2017image,shang2024lafa,zhang2024mixture} require joint training, architectural modifications, and often treat all images equally regardless of relation relevance.

In this work, we propose \textbf{FICHAD} (Fusion of Image Context and Human-Annotated Descriptions), a general framework for enabling structured visual reasoning in MMKGC. FICHAD synthesizes multimodal context—including link-aware descriptions, entity-centric summaries, and conceptual hints—from relevant images using pretrained VLMs. These outputs are integrated as textual inputs into existing description-based KGC models without modifying model architecture or requiring retraining.

FICHAD includes mechanisms to extract triple-relevant visual information and structure it in a way that aligns with the semantics of the query triple. This allows fine-grained reasoning over both symbolic and visual cues while maintaining compatibility with strong textual KGC baselines.

We evaluate FICHAD on three public MMKG benchmarks (FB15K-237-IMG, MKG-W, MKG-Y) using SimKGC and CSProm-KG, demonstrating consistent improvements—up to +4.6\% MRR and +3.5\% Hits@1 over prior state-of-the-art. Ablation studies further show that link-aware generation and conceptual hints offer significant gains for relation-specific reasoning.

\textbf{Our contributions are as follows:}
\begin{itemize}
    \item We introduce \textbf{FICHAD}, a general-purpose framework for MMKGC that enables structured visual reasoning by transforming multimodal evidence into textual context consumable by off-the-shelf description-based KGC models—without retraining or architecture modification.

    \item We develop mechanisms for relation-aware image filtering and multimodal context synthesis, supporting fine-grained reasoning through link-aware descriptions, conceptual hints, and entity-centric summaries.

    \item We demonstrate strong empirical results across three MMKG datasets and two model backbones, improving MRR by up to +4.6\% and Hits@1 by up to +3.5\%, while offering modularity, interpretability, and scalability.

\end{itemize}

\noindent Code and datasets are available at \url{https://github.com/mhd0528/LLM_MMKGC-pub}.

\section{RELATED WORK}
\paragraph{Knowledge Graph Completion}  
Traditional KGC models such as TransE~\cite{bordes2013translating}, RotatE~\cite{sun2018rotate}, and ComplEx~\cite{trouillon2016complex} embed entities and relations into continuous vector spaces to capture structural patterns. More recently, \textit{description-based} approaches like SimKGC~\cite{wang-etal-2022-simkgc} and CSProm-KG~\cite{chen-etal-2023-dipping} utilize textual descriptions to enhance generalization, especially for unseen or sparsely connected entities. However, these models operate purely on textual input and cannot incorporate visual signals.

\paragraph{LLMs for KGC}  
Large language models have been applied to KGC by reformulating it as structured question answering or context expansion tasks. MPIKGC~\cite{xu2024multi} generates entity and relation summaries to guide prediction, while KICGPT~\cite{Wei_2023} treats KGC as QA over structured prompts. Although these methods benefit from the reasoning capacity of LLMs, they are constrained to unimodal settings and face practical challenges such as high inference cost and limited multimodal integration.

\paragraph{Multimodal KGC}  
Multimodal KGC (MMKGC) integrates visual data with graph structure to enhance entity representation and link prediction. Early methods like IKRL~\cite{xie2017image} and TransAE~\cite{wang2019multimodal} incorporate image features into KGE pipelines. Recent models such as LAFA~\cite{shang2024lafa}, MoMoK~\cite{zhang2024mixture}, and MMKGR~\cite{zheng2023mmkgr} leverage modality-aware fusion or neighborhood aggregation but require retraining and custom architectures. MR-MKG~\cite{lee2024multimodal} aligns MMKGs with LLMs for QA and analogy tasks, but does not address link prediction. In contrast, our approach is architecture-agnostic and requires no retraining: it injects VLM-derived context directly into existing textual KGC models.

\paragraph{Vision-Language Models (VLMs)}  
VLMs such as CLIP~\cite{radford2021learning}, InstructBLIP~\cite{dai2023instructblip}, and QWEN2-VL~\cite{Qwen2VL} achieve strong performance in vision-language tasks like captioning, retrieval, and grounding. Despite their semantic richness, VLMs remain underutilized in KGC due to representational mismatch with structured models. We bridge this gap by transforming VLM outputs into structured, relation-aware textual context that supports symbolic reasoning—without requiring visual feature fusion or model finetuning.

\section{PRELIMINARIES} 
\subsection{Multimodal Knowledge Graphs (MMKGs)}
Multimodal knowledge graphs (MMKGs) extend traditional knowledge graphs by associating entities with multimodal information. Formally, an MMKG is defined as $\mathcal{G}_{MM} = \{(h, r_k, t, I_{h}, I_{t})\}$, where $I_{h}$ and $I_{t}$ are the sets of multimodal data (e.g., images) associated with the head entity $h$ and tail entity $t$, respectively. The structural information in MMKGs is represented by triples $(h, r_k, t) \subset \mathcal{E} \times \mathcal{R} \times \mathcal{E}$, similar to traditional KGs, while the multimodal data $I_{h}$ and $I_{t}$ provide additional context for the entities.


\subsection{Multimodal Knowledge Graph Completion (MMKGC)}
The goal of MMKGC is to predict missing links in MMKGs by leveraging both structural and multimodal information. Given a query $(h, r_k, ?)$ or $(?, r_k, t)$, the objective is to find the most plausible tail entity $t$ or head entity $h$ to complete the triple, while also utilizing multimodal data $I_{h}$ and $I_{t}$ to enhance predictions. 

Similar to KGC, MMKGC relies on a scoring function $f: \mathcal{E} \times \mathcal{R} \times \mathcal{E} \rightarrow \mathbb{R}$ that assigns a plausibility score to each candidate triple $(h, r_k, t)$. However, MMKGC extends this function to incorporate multimodal context:
\begin{equation}
    s(h, r_k, t | I_{h}, I_{t}) = f(h, r_k, t) + g(I_{h}, I_{t}),
\end{equation}
where $g(I_{h}, I_{t})$ models the contribution of multimodal data to the overall plausibility score.

\section{METHODOLOGY}
\label{sec:model}
\begin{figure*}
  \centering
  \includegraphics[width=\textwidth]{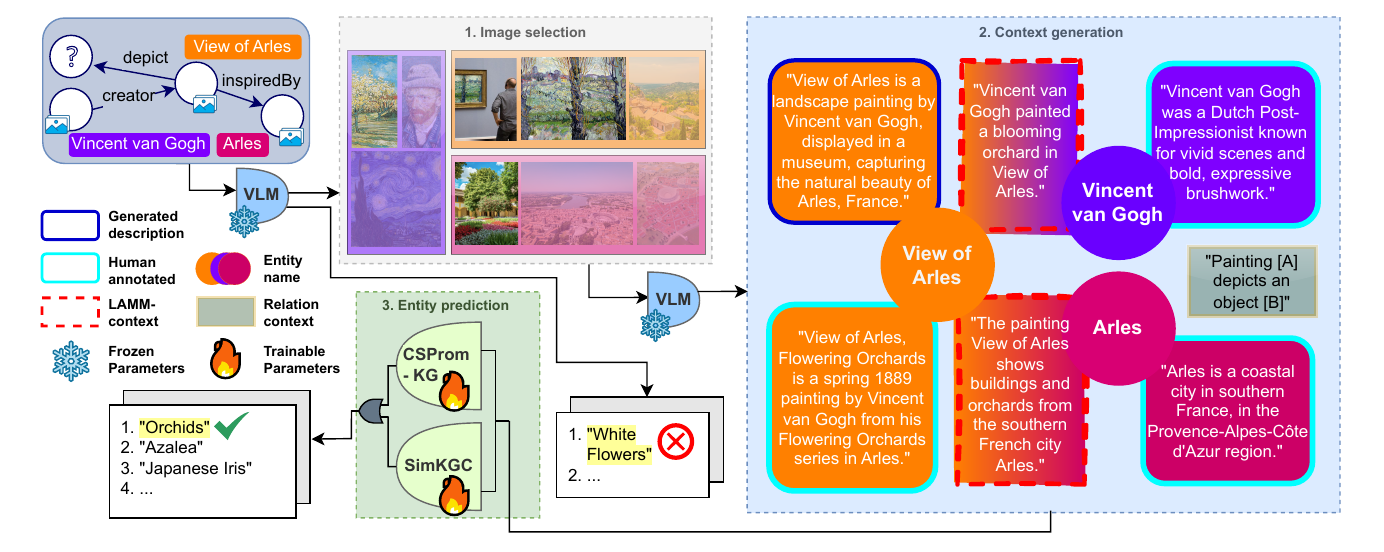}
  \caption{FICHAD pipeline. We generate multimodal context using pre-trained VLMs, either link-aware (filtered per relation) or entity-centric (based on all images per entity). Optional human-annotated descriptions and structural or relational templates are also integrated into the final prompt.}
  \setlength{\belowcaptionskip}{-20mm}
  \label{fig:pipeline}
\end{figure*}

\subsection{Overview}
\label{sec:Overview}
We propose \textbf{FICHAD}, a modular framework for enhancing description-based knowledge graph completion (KGC) by injecting structured multimodal context generated from pre-trained vision-language models (VLMs) as shown in Fig~\ref{fig:pipeline}. Rather than modifying KGC architectures to ingest image features, FICHAD transforms visual evidence into natural language descriptions, enabling symbolic reasoning with visual grounding.

Given a query triple $(h, r_k, ?)$, FICHAD proceeds in three steps:
\begin{enumerate}
    \item \textbf{Image selection:} Identify and filter images associated with $h$ and candidate $t$ based on their relevance to $r_k$.
    \item \textbf{Context generation:} Use a VLM to convert selected images into entity-specific or relation-aware textual descriptions.
    \item \textbf{Prompt construction:} Integrate the generated descriptions with structural neighbors and natural language relation templates to form the final input to a description-based KGC model.
\end{enumerate}

\noindent\textbf{Variants.} FICHAD supports four multimodal context configurations:
\begin{itemize}
    \item \textbf{FICHAD-1:} Link-aware descriptions based on filtered images of both $h$ and $t$.
    \item \textbf{FICHAD-2:} Entity-centric summaries using all images of $h$ or $t$ (no filtering).
    \item \textbf{FICHAD-1+x:} FICHAD-1 combined with human-annotated descriptions (e.g., from \texttt{DBpedia}).
    \item \textbf{FICHAD-1+y:} FICHAD-1 extended with conceptual hints to constrain the semantic type of the missing entity.
\end{itemize}

All variants share a unified input interface and are compatible with off-the-shelf KGC models (e.g., SimKGC~\cite{wang-etal-2022-simkgc}, CSProm-KG~\cite{chen-etal-2023-dipping}) without requiring retraining or architecture changes.

\subsection{FICHAD-1: Link-Aware Multimodal Context}
\label{sec:FICHAD-1}
FICHAD-1 constructs triple-specific context by identifying images relevant to both subject and object entities and summarizing them in relation-aware language.

\textbf{Image Filtering.} For each candidate image $\iota_j$, Qwen2-VL assigns a confidence score based on its alignment with the entity pair $(h, t)$:
\[
p(h, t, \iota_j) = \text{VLM}(\textit{prompt}_1(h, t), \iota_j),
\quad \text{retain if } p \geq \tau_r.
\]
For example, in the triple \texttt{(View of Arles, creator, Vincent van Gogh)}, images from the same series may be retained, while unrelated content like \textit{Starry Night} is filtered out.

\textbf{Entity Descriptions.} We generate concise visual summaries for $h$ and $t$:
\[
d_h = \text{VLM}(\textit{prompt}_2(h), I_h^*),
\quad d_t = \text{VLM}(\textit{prompt}_2(t), I_t^*).
\]
These capture visual and contextual features grounded in filtered images. For example:  
\textit{“View of Arles is an oil painting that depicts Arles, France using bold colors.”}

\textbf{Relation-Aware Summary.} A final sentence summarizes the visual co-occurrence and implicit relation:
\[
c_{h,t} = \text{VLM}(\textit{prompt}_3(h, t, d_h, d_t), I_h^*, I_t^*).
\]
E.g., for \texttt{(View of Arles, inspiredBy, Arles)}:  
\textit{“The painting View of Arles shows buildings and orchards from the southern French city Arles.”}

\subsection{FICHAD-2: Entity-Centric Context}
\label{sec:FICHAD-2}
FICHAD-2 offers broader, relation-agnostic visual descriptions by summarizing all images linked to an entity, without filtering. This is especially useful when few relation-relevant images are available.

Given image set $I_h$, the VLM generates:
\[
c_h = \text{VLM}(\textit{prompt}_4(h), I_h).
\]
Example:  
\textit{“View of Arles is a landscape painting by Van Gogh, capturing scenery in Arles, France.”}

These holistic descriptions enhance entity understanding and can serve as a fallback when FICHAD-1 is not applicable.

\subsection{Multimodal and Structural Context Integration}
\label{sec:context_integration}
To support richer reasoning, we include:
\begin{itemize}
    \item \textbf{Neighbor Contexts:} For each query entity, we retrieve $k$ neighbors and generate their FICHAD-based descriptions. These are appended to the final input prompt.
    \item \textbf{Relation Templates:} For each relation, we use Qwen2-VL to synthesize a human-readable template (e.g., \texttt{[A] is the creator of [B]}) from representative triples and associated images.
\end{itemize}

This combined context helps disambiguate candidate entities and injects both symbolic and visual knowledge into the reasoning process.

\subsection{Prompt Construction and Inference}
\label{sec:final_prompt}
We construct a structured input for the KGC model consisting of:
\begin{enumerate}
    \item The query triple $(h, r_k, ?)$ or $(?, r_k, t)$.
    \item Multimodal description for the query entity.
    \item Multimodal descriptions for $k$ neighbors (via FICHAD-1 or FICHAD-2).
    \item A relation template for $r_k$.
\end{enumerate}

This input is processed by a description-based KGC model to predict the missing entity. Since all content is textual, the model architecture remains unchanged and compatible with off-the-shelf systems.

\vspace{1mm}
\noindent An example prompt is shown in Code~\ref{lst:example_prompt}:
\begin{lstlisting}[style=custompython, caption={KGC Input for Prediction (View of Arles, depict, ?)}, label=lst:example_prompt]
Entity: View of Arles

# Generated Entity Description: 
View of Arles is a landscape painting by Vincent van Gogh, displayed in a museum, capturing the natural beauty of Arles, France...

# Neighbor Contexts:
# FICHAD-1
creator|Vicent van Gogh: 
Vincent van Gogh painted a blooming orchard in View of Arles.
inspiredBy|Arles: 
The painting View of Arles shows buildings and orchards from the southern French city Arles.
...
# FICHAD-2
creator|Vicent van Gogh: 
Vincent van Gogh was a artist known for his self-portrait and vivid paintings like Starry Night.
inspiredBy|Arles: 
Arles is a coastal city in southern France with colorful nature views and traditional buildings.
...

Relation: depict
# Relation Template: 
Painting [A] depict object [B].

Query: (View of Arles, depict, ?)
\end{lstlisting}

\section{Experiments}
\label{sec:Experiments}
\begin{table}[h!]
\centering
\begin{tabular}{S|SSS} \toprule
    {\textbf{Dataset}} & {\textbf{FB15K-237-IMG}} ~\cite{toutanova-chen-2015-observed} & {\textbf{MKG-W}} ~\cite{xu2022relationenhanced} & {\textbf{MKG-Y}} ~\cite{xu2022relationenhanced} \\ \midrule
    {\textbf{\#Entity}}  & {14,541} & {15,000} & {15,000} \\
    {\textbf{\#Relation}}  & {237} & {169} & {28} \\ 
    {\textbf{\#Images}}  & {179,341} & {40,771} & {46,279} \\ \midrule
    {\textbf{\#Train}}  & {272,115} & {34,196} & {21,310} \\ 
    {\textbf{\#Valid}}  & {17,535} & {4,276} & {2,665} \\
    {\textbf{\#Test}}  & {20,466} & {4,274} & {2,663} \\ \bottomrule
\end{tabular}
\caption{Dataset statistics}
\label{table:benchmarks}
\end{table}
\vspace{-5mm}

\subsection{Experimental Setup}

\paragraph{Datasets} We evaluate on three MMKGC benchmarks: FB15K-237-IMG~\cite{toutanova-chen-2015-observed}, MKG-W, and MKG-Y~\cite{xu2022relationenhanced}. Each dataset includes structural triples, database-derived descriptions (from Freebase or DBpedia), and entity-associated images. Table~\ref{table:benchmarks} summarizes their statistics.

\paragraph{Evaluation Protocol} We follow the standard link prediction setting~\cite{bordes2013translating}, reporting Mean Reciprocal Rank (MRR) and Hits@K (K=1, 3, 10) for head and tail prediction under the filtered evaluation protocol.

\paragraph{Models and Setup} We use SimKGC~\cite{wang-etal-2022-simkgc} and CSProm-KG~\cite{chen-etal-2023-dipping}, both based on BERT-base encoders. Models are trained on a single NVIDIA A100 GPU. For efficiency, CSProm-KG input is truncated to 120 tokens. Relation templates are used in all variants. Model details appear in Table~\ref{tab:model_settings}.

Multimodal context is generated using Qwen2-VL-7B with temperature 1.0. Images are filtered using link-aware prompts; We leverage the generation probability of token \texttt{yes} by the VLM as the relevancy score for each image; as the those scoring above $\tau_r = 0.85$ are grouped (max 5) to generate entity or relation-aware summaries. All context is generated offline.

\begin{table}[htbp]
\large
\centering
\caption{Model configurations and training details.}
\label{tab:model_settings}
\resizebox{\linewidth}{!}{%
\begin{tabular}{l|cc}
\toprule
\textbf{Component} & \textbf{SimKGC} & \textbf{CSProm-KG} \\
\midrule
\textbf{Entity Encoder} & BERT-base (uncased) & BERT-base (uncased)\\
\textbf{Input Sequence Length} & 50 tokens & 120 tokens\footnotemark \\
\textbf{Training Epochs (FB15k-237)} & 10 & 60 \\
\textbf{Training Epochs (MKG-W/Y)} & 10 & 200 \\
\textbf{Neighbors per Entity} & 5 & 8 \\
\midrule
\textbf{Images per Entity} & \multicolumn{2}{c}{10 (FB15k), 3 (MKG-W/Y)} \\
\textbf{Image Relevance Threshold $\tau_r$} & \multicolumn{2}{c}{0.85 (FICHAD-1)} \\
\bottomrule
\end{tabular}
}
\end{table}
\footnotetext{CSPromKG, by default, didn't have a constraint on the \# of tokens. We set \# of tokens to 120 due to efficiency concerns.}
\vspace{-5mm}

\subsection{Baselines}

\paragraph{Structural KGC Models} We include traditional structure-only models: TransE~\cite{bordes2013translating}, DistMult~\cite{yang2014embedding}, ComplEx~\cite{trouillon2016complex}, and RotatE~\cite{sun2018rotate}, which embed entities and relations without using text or images.

\paragraph{Multimodal KGC Models} We compare with MMKGC models that incorporate visual and/or textual signals: IKRL~\cite{xie2017image}, RSME~\cite{wang2021visual}, TransAE~\cite{wang2019multimodal}, OTKGE~\cite{cao2022otkge}, IMF~\cite{li2023imf}, VISTA~\cite{lee-etal-2023-vista}, NATIVE~\cite{DBLP:conf/sigir/ZhangCGXHLZC24}, and AdaMF-MAT~\cite{zhang-etal-2024-unleashing}. These use various fusion or alignment mechanisms to inject visual features into graph encoders.

\paragraph{Description-Based KGC Models} We focus on SimKGC and CSProm-KG as primary baselines, which take natural language descriptions as input and are directly compatible with our generated multimodal context. SimKGC appends entity descriptions to its input, while CSProm-KG uses them to retrieve contrastive soft prompts. These baselines also serve as our model backbones in all FICHAD variants.

\subsection{Our Model: FICHAD Variants}
\label{sec:our-model}

We introduce \textbf{FICHAD}, a modular approach that enhances description-based KGC with vision-language model (VLM) generated context. All FICHAD variants rely on Qwen2-VL to produce structured natural language descriptions based on entity-associated images and prompts tailored to relation semantics. This context is prepended to the KGC input without modifying model architecture.

\noindent\textbf{FICHAD-1 (Link-Aware Multimodal Context)}  
This variant filters images based on whether they jointly depict the head and tail entities, then generates a one-sentence description capturing the link-aware multimodal (LAMM) information for the triple.

\noindent\textbf{FICHAD-1+x (LAMM-Context + Database Description)}  
We concatenate the LAMM-context from FICHAD-1 with the first sentence of the entity’s human-annotated description, combining visual semantics and factual knowledge.

\noindent\textbf{FICHAD-1+y (LAMM-Context + Conceptual Hint)}  
We augment the LAMM-context with a conceptual hint that constrains the expected type of the missing entity based on the query relation and visual context. It serves as a scalable alternative when symbolic descriptions are unavailable.

\begin{table*}[ht]
\tiny
\centering
\caption{Comparison of FICHAD and baselines on FB15K-237-IMG, MKG-W, and MKG-Y datasets. $\pm$x indicates the absence or presence of Setting-x. The best results for each dataset are highlighted in bold. The second best with underline. The baseline results are taken from ~\cite{zhang2022knowledgegraph}, ~\cite{DBLP:conf/sigir/ZhangCGXHLZC24}, and ~\cite{lee-etal-2023-vista}.}
\label{tab:main_results}
\resizebox{\textwidth}{!}{
\begin{tabular}{l|cccc|cc|cc}
\toprule
\textbf{Model} & \multicolumn{4}{c}{\textbf{FB15K-237-IMG}} & \multicolumn{2}{c}{\textbf{MKG-W}} & \multicolumn{2}{c}{\textbf{MKG-Y}} \\
               & \textbf{MRR}         & \textbf{Hits@1}     & \textbf{Hits@3}     & \textbf{Hits@10}    & \textbf{MRR}         & \textbf{Hits@1}     & \textbf{MRR}         & \textbf{Hits@1}     \\
\midrule
\textbf{TransE}~\cite{bordes2013translating}                & 0.261       & 0.173      & 0.291      & 0.437      & 0.291       & 0.210      & 0.307       & 0.234       \\
\textbf{DistMult}~\cite{yang2014embedding}                  & 0.241       & 0.155      & 0.263      & 0.419      & 0.209       & 0.159      & 0.250       & 0.193       \\
\textbf{ComplEx}~\cite{trouillon2016complex}                & 0.247       & 0.158      & 0.275      & 0.428      & 0.249       & 0.190      & 0.287       & 0.222       \\
\textbf{RotatE}~\cite{sun2018rotate}                        & 0.338       & 0.241      & 0.375      & 0.533      & 0.337       & 0.268      & 0.325       & 0.291       \\ 
\textbf{PaiRE}~\cite{chao-etal-2021-pairre}                 & 0.351       & 0.256      & 0.387      & 0.544      & 0.344       & 0.282      & 0.320       & 0.255       \\ \midrule
\textbf{IKRL}~\cite{xie2017image}                           & 0.268       & 0.177      & 0.301      & 0.449      & 0.323       & 0.261      & 0.332       & 0.303       \\
\textbf{RSME}~\cite{wang2021visual}                         & 0.344       & 0.242      & 0.344      & 0.467      & 0.292       & 0.233      & 0.344       & 0.317       \\
\textbf{TransAE}~\cite{wang2019multimodal}                  & 0.211       & 0.199      & 0.317      & 0.463      & 0.300       & 0.212      & 0.281       & 0.253       \\
\textbf{OTKGE}~\cite{cao2022otkge}                    & 0.341       & 0.251      & 0.373      & 0.519      & 0.343       & 0.288      & 0.355       & 0.319       \\
\textbf{IMF}~\cite{li2023imf}                      & 0.367       & 0.273      & 0.404      & 0.557      & 0.345       & 0.287      & 0.357       & 0.329       \\
\textbf{VISTA}~\cite{lee-etal-2023-vista}                   & \textbf{0.380}       & \textbf{0.287}      & \textbf{0.415}      & \textbf{0.571}      & 0.329       & 0.261      & 0.304       & 0.248       \\
\textbf{NATIVE}~\cite{DBLP:conf/sigir/ZhangCGXHLZC24}       & -           & -          & -          & -          & 0.365       & 0.295      & 0.390       & 0.347       \\
\textbf{AdaMF-MAT}~\cite{zhang-etal-2024-unleashing}        & -           & -          & -          & -          & 0.358       & 0.290      & 0.385       & 0.343       \\ \midrule
\textbf{CSProm-KG (\texttt{-}x)}~\cite{chen-etal-2023-dipping}                            & 0.299       & 0.216      & 0.326      & 0.465      & 0.315       & 0.262      & 0.271       & 0.242       \\
\hspace{3mm}\textbf{\texttt{+} x} (original)                & 0.357       & 0.269      & 0.390      & 0.535      & 0.344       & 0.255      & 0.338       & 0.301       \\
\hspace{3mm}\textbf{\texttt{+} FICHAD-1}                     & 0.298       & 0.212      & 0.316      & 0.461      & 0.279       & 0.235      & 0.315       & 0.282       \\
\hspace{3mm}\textbf{\texttt{+} FICHAD-1 \texttt{+} x} (Ours) & \underline{0.368}       & \underline{0.275}      & \underline{0.406}      & \underline{0.558}      & 0.333       & 0.284      & 0.348       & 0.302       \\
\midrule
\textbf{SimKGC (\texttt{-}x)}~\cite{wang-etal-2022-simkgc}                               & 0.310       & 0.224      & 0.333      & 0.483      & 0.301       & 0.238      & 0.297       & 0.239       \\
\hspace{3mm}\textbf{\texttt{+} x} (original)                & 0.336       & 0.250      & 0.362      & 0.511      & \textbf{0.409}       & \textbf{0.325}      & \textbf{0.446}       & \textbf{0.369}       \\
\hspace{3mm}\textbf{\texttt{+} FICHAD-1}                     & 0.306       & 0.222      & 0.327      & 0.477      & 0.345       & 0.274      & 0.354       & 0.300       \\
\hspace{3mm}\textbf{\texttt{+} FICHAD-1 \texttt{+} x} (Ours) & 0.339       & 0.269      & 0.372      & 0.521      & \underline{0.392}       & \underline{0.312}      & \underline{0.426}       & \underline{0.353}       \\
\bottomrule
\end{tabular}
}
\vspace{2mm} 
\end{table*}

\subsection{Main Results}
\label{sec:main-results}

Table~\ref{tab:main_results} presents the performance of our FICHAD variants compared to existing KGC and MMKGC models across all three datasets.

FICHAD-1+x consistently outperforms prior methods. On FB15K-237-IMG, CSProm-KG + FICHAD-1+x achieves an MRR of 0.368, surpassing all baselines and approaching the top-performing VISTA model on Hits@K. On MKG-W and MKG-Y, SimKGC + FICHAD-1+x achieves MRRs of 0.392 and 0.426, improving over state-of-the-art methods like NATIVE and AdaMF-MAT by 3–4\%. These results show that combining structured visual context with symbolic KB descriptions leads to stronger entity and relation representations.

FICHAD-1 also shows strong performance without relying on human-annotated descriptions. Compared to the “–x” versions of SimKGC and CSProm-KG, FICHAD-1 improves MRR and Hits@1 by 3–5\% on MKG-W and MKG-Y, and remains competitive on FB15K-237-IMG. This confirms that our link-aware visual descriptions, even when used alone, provide useful semantic grounding for KGC models.

Notably, the combination of human-annotated and VLM-generated context in FICHAD-1+x yields the best results. While SimKGC directly integrates descriptions into its input and CSProm-KG uses them for retrieval, both benefit from the added multimodal context. This suggests that symbolic and visual signals offer complementary value in enhancing reasoning over the knowledge graph.

Finally, FICHAD remains effective across datasets with different image densities. While FB15K-237-IMG provides around 10 images per entity, MKG-W and MKG-Y offer only 3. Despite this, FICHAD-1+x performs robustly, indicating that VLMs like Qwen2-VL can extract meaningful context even in sparse visual settings.

In summary, FICHAD significantly improves MMKGC performance using a simple, modular approach that requires no architecture changes or VLM finetuning. Unlike fusion-based models, FICHAD incurs no training cost for multimodal integration, making it practical for plug-and-play deployment. It offers a practical and scalable alternative to existing embedding-based fusion techniques.

\begin{table*}[h]
\tiny
\centering
\caption{Ablation study results for FICHAD on MM description and Conceptual Hints. The best results of each model for each dataset are highlighted in bold. The second best are underlined.}
\label{tab:ablation_results}
\resizebox{\textwidth}{!}{
    \begin{tabular}{l|cccc|cc|cc}
    \toprule
    \textbf{Model} & \multicolumn{4}{c}{\textbf{FB15K-237-IMG}} & \multicolumn{2}{c}{\textbf{MKG-W}} & \multicolumn{2}{c}{\textbf{MKG-Y}} \\
                   & \textbf{MRR}         & \textbf{Hits@1}     & \textbf{Hits@3}     & \textbf{Hits@10}    & \textbf{MRR}         & \textbf{Hits@1}     & \textbf{MRR}         & \textbf{Hits@1}     \\
    \midrule
    \textbf{CSProm-KG }                         & 0.299            & 0.216            & 0.326            & 0.465            & 0.315            & 0.262            & 0.271            & 0.242         \\
    \hspace{3mm}\textbf{\texttt{+} x}           & \textbf{0.357}   & \textbf{0.269}   & \textbf{0.390}  & \textbf{0.535}   & \textbf{0.344}   & \underline{0.255}& \textbf{0.338}   & \textbf{0.301}         \\
    \hspace{3mm}\textbf{\texttt{+} FICHAD-1}     & 0.298            & 0.212            & 0.316            & 0.461            & 0.279            & 0.235            & \underline{0.315}& \underline{0.282}         \\
    \hspace{3mm}\textbf{\texttt{+} FICHAD-2}     & 0.268            & 0.190            & 0.292            & 0.422            & 0.250            & 0.216            & 0.300            & 0.274         \\
    \hspace{3mm}\textbf{\texttt{+} FICHAD-1+y}   & \underline{0.348}& \underline{0.252}& \underline{0.376}& \underline{0.521}& \underline{0.329}& \textbf{0.274}  & 0.242            & 0.214         \\
    \midrule
    \textbf{SimKGC (\texttt{-}x)}               & 0.310            & 0.224            & 0.333            & 0.483            & 0.301            & 0.238            & 0.297            & 0.239         \\
    \hspace{3mm}\textbf{\texttt{+} x}           & \textbf{0.336}   & \textbf{0.250}  & \underline{0.362}& \underline{0.511}& \textbf{0.409}   & \textbf{0.325}   & \textbf{0.446}   & \textbf{0.369}       \\
    \hspace{3mm}\textbf{\texttt{+} FICHAD-1}     & 0.306            & 0.222            & 0.327            & 0.477            & \underline{0.345}& \underline{0.274}& 0.354            & 0.300         \\
    \hspace{3mm}\textbf{\texttt{+} FICHAD-2}     & 0.234            & 0.161            & 0.250            & 0.380            & 0.299            & 0.230            & 0.257            & 0.200         \\
    \hspace{3mm}\textbf{\texttt{+} FICHAD-1+y}   & \underline{0.327}& \underline{0.247}& \textbf{0.367}   & \textbf{0.512}  & 0.326            & 0.245            & \underline{0.434}& \underline{0.380} \\
    \bottomrule
    \end{tabular}
    }
\end{table*}

\begin{table}[]
\tiny
\centering
\caption{The table below presents entity context statistics across FB15K-237-IMG, MKG-W, and MKG-Y.}
\label{tab:context_stats}
\resizebox{\columnwidth}{!}{
    \begin{tabular}{l|c|c|c|c} 
    \toprule
    \multirow{2}{*}{\textbf{Dataset}} & \multirow{2}{*}{\textbf{\#Entity}} & \multicolumn{3}{c}{\textbf{\#Entity with}} \\
    & & \textbf{Images} & \textbf{FICHAD-1} & \textbf{FICHAD-2} \\
    \midrule
    \textbf{FB15K-237-IMG}  & {14,541}  & {14,541}  & {14,541}  & {14,541}  \\
    \textbf{MKG-W}          & {15,000}  & {14,463}  & {10,318}  & {14,463}  \\
    \textbf{MKG-Y}          & {15,000}  & {14,244}  & {10,712}  & {14,244}  \\
    \bottomrule
\end{tabular}
}
\end{table}

\subsection{Ablation of Link-Aware Generation}
We first evaluate the impact of link-awareness in multimodal context generation by comparing two VLM-generated variants: FICHAD-1 and FICHAD-2. FICHAD-2 summarizes all available images for each entity independently of its relationships, while FICHAD-1 filters link-relevant images and conditions generation on both head and tail entities.

As shown in Table~\ref{tab:ablation_results}, FICHAD-1 consistently outperforms FICHAD-2 across all datasets and metrics. For instance, SimKGC’s MRR improves from 0.299 to 0.345 on MKG-W, and Hits@1 improves from 0.200 to 0.300 on MKG-Y. These gains demonstrate the benefit of incorporating relational cues and filtering irrelevant images.

The performance gap underscores the limitations of entity-centric generation: without explicit grounding, VLMs often focus on incidental attributes rather than relation-relevant features. Link-aware descriptions better capture triple semantics and support more accurate KGC reasoning. This validates our strategy of converting visual information into structured text for models that cannot process raw visual inputs. In the next subsection, we compare FICHAD-2 to Setting-x to examine differences in modality and data source quality.

\subsection{Ablation of Modality and Data Sources}
We next compare FICHAD-2 with Setting-x, which uses database-derived entity descriptions (e.g., from Freebase or DBpedia). While both enhance entity understanding, FICHAD-2 relies on VLM-generated descriptions from images, while Setting-x uses manually curated textual facts.

As shown in Table~\ref{tab:ablation_results}, Setting-x consistently outperforms FICHAD-2. On MKG-W, SimKGC achieves MRR 0.409 with Setting-x compared to 0.299 with FICHAD-2. On MKG-Y, Hits@1 improves from 0.200 (FICHAD-2) to 0.369 (Setting-x). These results reflect the strength of well-curated text and the challenges of image-based abstraction.

Two factors contribute to the gap: (1) entity images may include noise or co-occurring objects that distract the VLM, and (2) current VLMs are more effective at describing visual scenes than distilling KG-relevant facts. As a result, FICHAD-2 often produces verbose or imprecise outputs. Still, it serves as a valuable fallback when human-annotated descriptions are missing and represents a step toward multimodal KGC with minimal supervision.

\subsection{Impact of Conceptual Hints on Query Relation Understanding}
We now assess whether conceptual hints improve query relation understanding and help constrain the prediction space. FICHAD-1+y extends FICHAD-1 by appending a conceptual hint generated from the query entity, relation, and filtered images.

For each relation, we gather triples from the KG with the same relation and feed them to an LLM. The LLM then summarizes the likely entity types for that relation. For example, in Code~\ref{lst:conceptual_hints_examples} with the relation (\texttt{depicts}), we collect 20 triples from the KG and ask the LLM to suggest possible entity types for the given entity (\texttt{View of Arles}), which is a painting. 

As shown in Table~\ref{tab:ablation_results}, FICHAD-1+y consistently improves over FICHAD-1 and approaches the performance of Setting-x. For example, on FB15K-237-IMG with SimKGC, FICHAD-1+y reaches MRR 0.327 and Hits@10 0.512—nearly matching Setting-x (0.336/0.511) and significantly outperforming FICHAD-1 alone (0.306/0.477).

The benefit comes from encoding relational expectations in natural language form. Unlike generic templates like \texttt{[A] depicts [B]}, conceptual hints include type-level priors grounded in visual semantics (for example Code~\ref{lst:conceptual_hints_examples}). These help the model narrow its candidate space and reinforce structural consistency. Importantly, FICHAD-1+y does not require any curated descriptions, offering a scalable and symbolic-free augmentation strategy for MMKGC.

\begin{lstlisting}[style=custompython, breaklines=true, breakatwhitespace=true, caption={Conceptual hint for (View of Arles, depict, ?)}, label=lst:conceptual_hints_examples]
Query: View of Arles, depict, ?
Relation template: [A] depicts [B]. 
Conceptual hint: 
    Question: What does View of Arles depict? 
    Hint: The missing entity is likely a physical object like a 'plant' or 'building', based on the context of the 'painting' View of Arles.
\end{lstlisting}

\begin{table}[]
\large
\centering
\caption{Context Quality Evaluation Across FB15k-237-IMG, MKG-W, and MKG-Y}
\label{table:quality_evaluation}
\resizebox{\columnwidth}{!}{
    \begin{tabular}{l|c|c|c|c} 
    \toprule
    \multirow{1}{*}{\textbf{Dataset}} & \multicolumn{2}{c}{\textbf{FICHAD-1}} & \multicolumn{2}{c}{\textbf{FICHAD-2}} \\
    & \makecell{\textbf{Single Entity}\\\textbf{Coverage}} & \makecell{\textbf{Both Entity}\\\textbf{Coverage}} & \makecell{\textbf{Entity}\\\textbf{Coverage}} & \makecell{\textbf{BERT}\\\textbf{Score}} \\
    \midrule
    \textbf{FB15K-237-IMG}  & {0.98}  & {0.71}  & {1.00}  & {0.83}  \\
    \textbf{MKG-W}          & {0.98}  & {0.70}  & {1.00}  & {0.84}  \\
    \textbf{MKG-Y}          & {0.98}  & {0.73}  & {1.00}  & {0.81}  \\
    \bottomrule
\end{tabular}
}
\end{table}

\subsection{Challenges in Multimodal Context Generation}
\label{sec:context_challenges}

FICHAD-1 and FICHAD-2 rely on vision-language models (VLMs) to generate structured textual context from images. While this enables multimodal reasoning without architectural changes, two main challenges affect the quality and consistency of generated descriptions: \textbf{visual sparsity} and \textbf{hallucination}.

\textbf{Visual sparsity.} Despite near-complete image coverage in FB15K-237-IMG, entity coverage in MKG-W and MKG-Y drops to 96.4\% and 94.9\% as shown in Table~\ref{tab:context_stats}. Furthermore, FICHAD-1’s link-aware filtering reduces usable triples because only images depicting both head and tail entities are retained. For instance, in MKG-W, only 10,318 triples retain link-aware context, compared to 14,463 in FICHAD-2. To maintain consistency across samples, we allow fallback generation based on the entity name alone when no images pass the filtering threshold. While this preserves format, it weakens visual grounding.

Examples of generated LAMM-contexts show that when sufficient images are available, descriptions capture both visual and commonsense information. When imagery is limited or ambiguous, the model either defaults to entity co-occurrence (e.g., "X is related to Y") or generates surface-level visual summaries (e.g., describing objects without linking them to the entity).

\textbf{Visual hallucination.} VLMs occasionally produce irrelevant or inaccurate outputs, especially when images contain cluttered scenes or co-occurring entities. This is more prevalent in FICHAD-2, where all images are used without filtering. In such cases, the generated description may list low-salience visual features or misattribute them, leading to hallucinated facts. By contrast, FICHAD-1 benefits from explicit link conditioning, which constrains the generation space and improves entity alignment.

To assess description quality, we report BERTScore~\cite{zhang2019bertscore} similarity between FICHAD-2 outputs and database-derived descriptions, and entity coverage for both variants (Table~\ref{table:quality_evaluation}). FICHAD-2 achieves 100\% coverage across all datasets, with BERTScores of 0.83 (FB15K), 0.84 (MKG-W), and 0.81 (MKG-Y). FICHAD-1 achieves 98\% coverage for at least one entity, and 70–73\% for both, due to stricter filtering.

These results highlight that while image-to-text conversion enables compatibility with language-based KGC models, its effectiveness is tied to image quality and selection. Nonetheless, both FICHAD variants deliver strong semantic grounding. These insights motivate future work in filtering image noise and calibrating VLM generation through symbolic constraints. Future work may further mitigate hallucination by incorporating retrieval-based filtering or multimodal consistency checks.

\section{Conclusion and future work}
\label{sec:future-work}
We presented \textbf{FICHAD}, a general framework for knowledge graph completion (KGC) that leverages structured, link-aware multimodal context generated by vision-language models (VLMs). Without requiring model architecture modifications or retraining, FICHAD introduces a lightweight preprocessing pipeline that converts relevant images into textual descriptions aligned with the semantics of each query triple.

Experiments on three MMKG benchmarks—FB15K-237-IMG, MKG-W, and MKG-Y—demonstrate that FICHAD significantly improves link prediction performance over existing multimodal KGC methods. In particular, the hybrid variant FICHAD-1+x, which combines VLM-generated context with database-derived descriptions, achieves state-of-the-art results across most metrics. We also show that conceptual hints can enhance query relation understanding by providing soft semantic constraints that guide the prediction space.

Beyond performance, our analysis of context quality using BERTScore and entity coverage highlights the scalability and utility of VLM-generated descriptions, even under sparse or noisy visual settings.

Future directions include extending FICHAD to incorporate richer modalities such as video or audio, improving the robustness of image selection and multimodal summarization, and applying FICHAD-style structured context to downstream tasks like multimodal question answering or retrieval-augmented generation. As a plug-and-play module, FICHAD offers a practical and scalable path to integrating visual signals into symbolic reasoning pipelines.

\bibliographystyle{splncs04}
\bibliography{ref}

\end{document}